\documentclass[11pt]{article}

\usepackage[a4paper,margin=2.5cm,heightrounded=true]{geometry}
\usepackage{times}
\usepackage{latexsym}
\usepackage[T1]{fontenc}
\usepackage[utf8]{inputenc}
\usepackage{microtype}
\usepackage{inconsolata}
\usepackage{graphicx}
\usepackage{booktabs}
\usepackage{array}
\usepackage{amsmath,amssymb}
\usepackage{mathtools}
\usepackage{natbib}
\usepackage{xcolor}
\usepackage[switch,mathlines]{lineno}
\usepackage{etoolbox}
\usepackage{caption}
\usepackage{url}
\usepackage[breaklinks]{hyperref}
\usepackage{placeins}
\usepackage{balance}

\graphicspath{{assets/}}
\twocolumn
\makeatletter

\newcount\cv@tmpc@ \newcount\cv@tmpc
\def\fillzeros[#1]#2{\cv@tmpc@=#2\relax\ifnum\cv@tmpc@<0\cv@tmpc@=-\cv@tmpc@\fi
  \cv@tmpc=1 %
  \loop\ifnum\cv@tmpc@<10 \else \divide\cv@tmpc@ by 10 \advance\cv@tmpc by 1 \fi
    \ifnum\cv@tmpc@=10\relax\cv@tmpc@=11\relax\fi \ifnum\cv@tmpc@>10 \repeat
  \ifnum#2<0\advance\cv@tmpc1\relax-\fi
  \loop\ifnum\cv@tmpc<#1\relax0\advance\cv@tmpc1\relax\fi \ifnum\cv@tmpc<#1 \repeat
  \cv@tmpc@=#2\relax\ifnum\cv@tmpc@<0\cv@tmpc@=-\cv@tmpc@\fi \relax\the\cv@tmpc@}%

\newcommand*\linenomathpatch[1]{%
  \expandafter\pretocmd\csname #1\endcsname {\linenomath}{}{}%
  \expandafter\pretocmd\csname #1*\endcsname {\linenomath}{}{}%
  \expandafter\apptocmd\csname end#1\endcsname {\endlinenomath}{}{}%
  \expandafter\apptocmd\csname end#1*\endcsname {\endlinenomath}{}{}%
}
\newcommand*\linenomathpatchAMS[1]{%
  \expandafter\pretocmd\csname #1\endcsname {\linenomathAMS}{}{}%
  \expandafter\pretocmd\csname #1*\endcsname {\linenomathAMS}{}{}%
  \expandafter\apptocmd\csname end#1\endcsname {\endlinenomath}{}{}%
  \expandafter\apptocmd\csname end#1*\endcsname {\endlinenomath}{}{}%
}
\expandafter\ifx\linenomath\linenomathWithnumbers
  \let\linenomathAMS\linenomathWithnumbers
  \patchcmd\linenomathAMS{\advance\postdisplaypenalty\linenopenalty}{}{}{}
\else
  \let\linenomathAMS\linenomathNonumbers
\fi
\AtBeginDocument{%
  \linenomathpatch{equation}%
  \linenomathpatchAMS{gather}%
  \linenomathpatchAMS{multline}%
  \linenomathpatchAMS{align}%
  \linenomathpatchAMS{alignat}%
  \linenomathpatchAMS{flalign}%
}

\newlength\titlebox
\renewcommand\maketitle{\par
 \begingroup
   \def\thefootnote{\fnsymbol{footnote}}
   \twocolumn[\@maketitle]
   \@thanks
 \endgroup
 \setcounter{footnote}{0}
 \let\maketitle\relax
 \let\@maketitle\relax
 \gdef\@thanks{}\gdef\@author{}\gdef\@title{}\let\thanks\relax}
\def\@maketitle{\vbox to \titlebox{\hsize\textwidth
 \linewidth\hsize \vskip 0.125in minus 0.125in \centering
 {\Large\bfseries \@title \par} \vskip 0.2in plus 1fil minus 0.1in
 {\large \@author \par}
 \vskip 0.3in plus 2fil minus 0.1in
}}

\renewenvironment{abstract}%
  {\begin{center}\large\textbf{\abstractname}\end{center}%
   \begin{list}{}{%
     \setlength{\rightmargin}{0.6cm}%
     \setlength{\leftmargin}{0.6cm}}%
   \item[]\ignorespaces%
   \@setsize\normalsize{12pt}\xpt\@xpt}%
  {\unskip\end{list}}

\def\section{\@startsection {section}{1}{\z@}{-2.0ex plus -0.5ex minus -.2ex}{1.5ex plus 0.3ex minus .2ex}{\large\bfseries\raggedright}}
\def\subsection{\@startsection{subsection}{2}{\z@}{-1.8ex plus -0.5ex minus -.2ex}{0.8ex plus .2ex}{\normalsize\bfseries\raggedright}}
\def\subsubsection{\@startsection{subsubsection}{3}{\z@}{-1.5ex plus -0.5ex minus -.2ex}{0.5ex plus .2ex}{\normalsize\bfseries\raggedright}}
\def\paragraph{\@startsection{paragraph}{4}{\z@}{1.5ex plus 0.5ex minus .2ex}{-1em}{\normalsize\bfseries}}
\def\subparagraph{\@startsection{subparagraph}{5}{\parindent}{1.5ex plus 0.5ex minus .2ex}{-1em}{\normalsize\bfseries}}

\def\thebibliography#1{\vskip\parskip%
\vskip\baselineskip%
\def\baselinestretch{1}%
\ifx\@currsize\normalsize\@normalsize\else\@currsize\fi%
\vskip-\parskip
\section*{References\@mkboth{References}{References}}\list
 {}{\setlength{\labelwidth}{0pt}\setlength{\leftmargin}{\parindent}%
 \setlength{\itemindent}{-\parindent}}%
 \def\newblock{\hskip .11em plus .33em minus -.07em}%
 \sloppy\clubpenalty4000\widowpenalty4000\sfcode`\.=1000\relax}

\labelwidth\leftmargini\advance\labelwidth-\labelsep \labelsep 5pt
\def\@listi{\leftmargin\leftmargini}
\def\@listii{\leftmargin\leftmarginii
 \labelwidth\leftmarginii\advance\labelwidth-\labelsep
 \topsep 2pt plus 1pt minus 0.5pt
 \parsep 1pt plus 0.5pt minus 0.5pt
 \itemsep \parsep}
\belowdisplayskip \abovedisplayskip
\def\@normalsize{\@setsize\normalsize{11pt}\xpt\@xpt}
\def\small{\@setsize\small{10pt}\ixpt\@ixpt}
\def\footnotesize{\@setsize\footnotesize{10pt}\ixpt\@ixpt}
\def\scriptsize{\@setsize\scriptsize{8pt}\viipt\@viipt}
\def\tiny{\@setsize\tiny{7pt}\vipt\@vipt}
\def\large{\@setsize\large{14pt}\xiipt\@xiipt}
\def\Large{\@setsize\Large{16pt}\xivpt\@xivpt}
\makeatother

\definecolor{darkblue}{rgb}{0,0,0.5}
\hypersetup{colorlinks=true,citecolor=darkblue,linkcolor=darkblue,urlcolor=darkblue}
\hypersetup{pdftitle={Who Judges Matters: Measuring Family-Conditioned Preference in LLM-as-Judge Panels},pdfauthor={David Ababio Awuni; Luke E. K. Achenie; Benjamin Tei Partey; Elvis Gyasi Owusu; Nii-Nai Derrick Sowah}}

\title{Who Judges Matters: Measuring Family-Conditioned Preference in LLM-as-Judge Panels}
\author{David Ababio Awuni \quad Luke E. K. Achenie\thanks{Corresponding author.} \quad Benjamin Tei Partey\\
Elvis Gyasi Owusu \quad Nii-Nai Derrick Sowah}
\date{}

\begin{document}
\maketitle

\begin{abstract}
Who the judge is can affect an LLM-as-judge result, but measuring that effect without confusing it with candidate quality is difficult. We study four open-weight families (Llama 3.1, Qwen 2.5, Gemma 2, and Yi 1.5) in a fully crossed pairwise design with 9,312 judgments. A common per-family statistic is strongly confounded with candidate quality and correlates with Bradley--Terry ability at $r=0.95$. We derive a corrected estimator that holds the candidate family fixed and compares judges. All four families then show a positive same-family lift (3.4--8.4 percentage points), with global FPS 0.067 (95\% CI [0.053, 0.084], permutation $p=0.0002$). The effect remains under panel-based quality controls, an independent human-consensus anchor, and a float16 judging replication. Judge-side likelihood is closely related to the effect: adding likelihood advantage reduces the controlled coefficient by 61\%, which we treat as descriptive attenuation rather than causal mediation. Position is a separate failure mode. Across the panel, 55.4\% of AB/BA pairs reverse, and reversal above 50\% is incompatible with a simple independent content-noise model. Relative to a family-balanced reference, panel composition changes 18.5\% of pairwise outcomes. A complete reproducibility archive has been prepared for public release.
\end{abstract}

\section{Introduction}
LLM judges are now used to rank models, select preference data, and support model-development decisions. This use assumes that, after ordinary quality differences are accounted for, changing the judge should not systematically change which response wins. A Qwen judge and a Yi judge will disagree on some examples; the concern is whether one judge family consistently gives extra support to one candidate family.

We test for that pattern in a fully crossed panel of open-weight models. We call it \emph{family-conditioned preference}: extra support for a candidate when the judge comes from the same model family. No judge evaluates its own generations, so this is not ordinary self-preference, and the setting is a fixed evaluation panel rather than a training pipeline with possible preference leakage.

Judge identity is already known to affect evaluation. LLM judges show position, verbosity, and self-enhancement biases \citep{zheng2023,wang2024,shi2025}; some favor their own outputs \citep{panickssery2024,liu2024,wataoka2024}. Family affinity has also appeared in pointwise and rubric-based settings \citep{spiliopoulou2025,pombal2026}, while functional similarity between judge and candidate can shift scores \citep{goel2025}. We therefore focus on three measurement questions rather than claiming to discover family-related preference itself.

\paragraph{First, how should family preference be measured in a zero-sum pairwise panel?}
The obvious statistic compares how much a judge supports its own family with how much it supports other candidate families. In a zero-sum pairwise panel, that comparison mixes candidate quality with the judge--candidate interaction. The problem is large in our data: the row-wise statistic correlates with Bradley--Terry ability at $r=0.95$. It makes Qwen look strongly biased ($+0.160$) and Yi almost unbiased ($-0.004$). We instead hold the candidate family fixed and compare judges: \emph{does the same candidate receive more support from its same-family judge than from the other judges?} This column-wise comparison removes the candidate main effect. The resulting lifts are positive for all four families, ranging from 0.034 to 0.084.

\paragraph{Second, does the effect survive quality controls?}
We next test whether measured quality differences explain the result. A nested fractional-logit analysis adds Bradley--Terry quality, style similarity, response length, candidate-side orientation, and family fixed effects, with little change in the same-family coefficient; a separate quasi-binomial GEE provides a correlation-structure robustness check. Because several controls come from the audited panel, we also use blind human consensus as an independent anchor; on the matched subset, adding the human label changes the coefficient from 0.370 to 0.382. A separate float16 judging run gives 0.066 globally, compared with 0.067 in the main quantized stack.

\paragraph{Third, what measurable property of the judge--candidate pair is associated with the effect?}
A plausible explanation is familiarity: judges may favor text they find easier to predict \citep{wataoka2024,stureborg2024,oi2024}. Same-family responses have a 0.69 natural-log-unit higher mean per-token log-probability than cross-family responses. Adding judge-side likelihood advantage to the fully controlled model reduces the same-family coefficient from 0.299 to 0.116, a 61\% attenuation, while likelihood advantage strongly predicts support ($\beta=0.144$, $p=7\times10^{-6}$). We treat this as descriptive rather than causal, and a smaller significant family effect remains.

Position instability is a different problem. Across the panel, 55.4\% of AB/BA pairs choose different winners, with Gemma at 78.8\%. A simple independent symmetric content-noise model cannot produce reversal above 50\%, so higher rates indicate systematic order dependence. Counterbalanced reconciliation cancels pure position-following, but reversed pairs become ties and any genuine family-preference signal on those pairs is lost. Position bias and family preference therefore need separate diagnostics.

The effect is also large enough to matter in practice. The 6.7 percentage-point global lift exceeds 8 of 9 adjacent length-controlled win-rate gaps in the AlpacaEval 2.0 top-10 snapshot used here, whose median adjacent gap is 1.09 pp. The exact reference table used for this comparison is included in the prepared reproducibility archive. This is only a scale comparison, not a claim that our panel reproduces or reranks AlpacaEval. Relative to a family-balanced reference, panel composition changes 18.5\% of pairwise outcomes.

\paragraph{Contributions.}
The paper contributes five pieces of evidence and methodology:
\begin{itemize}
    \item We show why the natural row-wise per-family contrast is confounded by candidate quality and give a column-wise estimator that isolates the judge--candidate family interaction.
    \item With the corrected estimator, every tested family has a positive and individually significant same-family lift. All four families also replicate in the positive direction under an independent float16 judging stack.
    \item We check quality using panel-derived controls, off-diagonal Bradley--Terry estimates, and an independent human-consensus anchor.
    \item We measure a likelihood-familiarity channel. Judge-side likelihood advantage is strongly associated with support and reduces the controlled same-family coefficient by 61\%, while a significant residual remains.
    \item We treat position instability as a separate problem and show why counterbalanced reconciliation removes position-following while also attenuating genuine preference signal on reversed pairs.
\end{itemize}

\section{Related Work}
\paragraph{LLM judges as measurement instruments.}
Early LLM-as-judge work reported high human agreement alongside position, verbosity, and self-enhancement biases \citep{zheng2023}. Later studies showed that response order can change pairwise rankings \citep{wang2024} and developed more direct position-bias diagnostics \citep{shi2025}. Recent work treats the judge itself as part of the measurement system that needs auditing \citep{yang2026b,usami2026}; JudgeArena similarly makes judges and backends swappable while retaining detailed metadata \citep{lushtaku2026}. Our fully crossed design follows this view by treating judge provenance as an experimental factor.

\paragraph{Self-preference, family preference, and quality confounding.}
Self-preference work asks whether a judge favors its own generations \citep{panickssery2024,liu2024,wataoka2024,yang2026a}; our judges never score their own outputs, so the relevant overlap is family-level. Quality adjustment is also central. \citet{chen2025reason} use verifiable tasks to separate harmful self-preference from genuinely better own-family outputs, and \citet{chen2025beyond} show that naive differences can mix bias with response quality. \citet{spiliopoulou2025} adjust for completion quality and report family bias, while \citet{pombal2026} find same-family preference in rubric-based evaluation. In our fully crossed zero-sum pairwise panel, this issue appears as a specific per-family identification error that can be corrected without a gold score for every response.

\paragraph{Similarity, familiarity, and likelihood.}
Similarity provides one possible explanation. \citet{goel2025} find that judges favor functionally similar models, extending the concern beyond exact self-preference. Other work connects self-preference to perplexity \citep{wataoka2024}, familiarity to evaluator inconsistency \citep{stureborg2024}, and likelihood to biased evaluation \citep{oi2024}. We therefore log judge-side token likelihood for each candidate response and measure how the controlled family coefficient changes when this signal is added.

\paragraph{Panels, correlated errors, and pipeline effects.}
Panels can reduce single-judge errors without eliminating dependence between judges. Diverse juries can outperform individual judges \citep{verga2024}, but correlated errors reduce effective panel diversity \citep{kohli2026}. Preference leakage creates a related provenance problem during training when evaluator and synthetic-data sources overlap \citep{li2026}. We study evaluation instead: how panel composition changes pairwise decisions relative to a family-balanced reference. Concurrent mitigation work likewise finds that reliability depends on the judge, prompt, and bias-control strategy \citep{yang2026a,soumik2026}.

\paragraph{Positioning.}
Prior work therefore establishes several routes through which evaluator provenance can matter. Our claim is narrower: in fully crossed zero-sum pairwise panels, we correct a per-family measurement error, show positive corrected lifts across all four tested families, relate much of the controlled coefficient to judge-side likelihood, and distinguish the result from position instability.

\section{Experimental Design}
\paragraph{Panel.}
The panel contains four open-weight families at two scales: Llama 3.1 (70B/8B), Qwen 2.5 (72B/7B), Gemma 2 (27B/9B), and Yi 1.5 (34B/9B). The main stack uses GGUF Q4\_K\_M models with \texttt{llama-cpp-python} and no proprietary API. Falcon 40B was piloted but excluded from the Primary-4 headline under criteria set before FPS was computed: its judge tie rate was 88.5\% versus a 3.4\% retained-panel mean, and its mean candidate support was 0.063 versus 0.56--0.65. Including Falcon would therefore mix a major response-health/quality failure with the preference quantity. Full-5 results appear only as specification-curve sensitivity evidence (Appendix~\ref{app:falcon}).

\paragraph{Prompts.}
A timestamped stratified split assigned 200 prompts to 134 exploratory and 66 confirmatory items before analysis. Six prompts, two exploratory and four confirmatory, failed the pre-specified response-health check because at least one candidate response did not meet the required length/quality condition. The primary set contains 194 prompts: 132 exploratory and 62 confirmatory, drawn from MT-Bench (78), AlpacaEval 2.0 (68), and WildBench (48) across 12 task categories \citep{zheng2023,dubois2024,lin2025}. Split and health-filter files are included in the release.

\paragraph{Fully crossed trials.}
Every judge family evaluates every unordered pair of candidate families under both response orders. The design contains
\[
194\ \text{prompts}\times 6\ \text{pairs}\times 2\ \text{orders}\times 4\ \text{judges}=9{,}312
\]
This produces 9{,}312 raw trials and 4,656 reconciled judge--prompt--pair observations. Each of the 16 judge$\times$candidate cells has exactly 582 observations. The matrix estimators therefore do not depend on unequal cell counts.

\paragraph{AB/BA reconciliation.}
For each focal candidate, an order-specific verdict is coded as $w\in\{0,0.5,1\}$, after which AB and BA are aligned to the same focal orientation and averaged. If the two orders name different winners, the reconciled score is $s=0.5$. Raw order dependence is substantial: the first-position A-rate is 0.778 and AB/BA agreement is 0.446. We therefore do not use single-presentation estimates as primary evidence; Section~\ref{sec:position} explains what reconciliation removes and what signal it can also discard.

\paragraph{Judge-side likelihood.}
For every trial, we compute the judge's teacher-forced mean per-token log-probability for each candidate response using the first 400 characters. This adds one forward pass per response and gives the familiarity measure used in Section~\ref{sec:familiarity}.

\paragraph{Human reference set.}
Three graduate-level NLP researchers labeled 400 blinded pairwise items as A/B/Tie with 1--5 confidence, stratified across all six candidate-family pairs with provenance hidden. One annotator was excluded before any FPS analysis under pre-specified quality criteria (Appendix~\ref{app:human}). The two retained annotators reached consensus on 266 items (Cohen's $\kappa=0.482$; 66.5\% exact agreement). On those items, the panel matches human consensus 57.1\% of the time, 15.0 pp above the 42.1\% majority-class baseline. We use these labels as an independent quality control and task-difficulty reference, not to define or validate FPS.

\section{Measuring Per-Family Preference Correctly}
\label{sec:measurement}
Let $P(j,c)$ be the mean reconciled support that judge family $j$ gives candidate family $c$. Every pairwise comparison distributes one unit of support across the two candidates. Because the candidate design is balanced, each judge row in $P$ sums to 2.00. That equality comes from the design rather than from the observed behavior.

We first define a global Family Preference Score (FPS) as diagonal support minus off-diagonal support:
{\small
\begin{equation}
\mathrm{FPS}=\frac{1}{|F|}\sum_f P(f,f)-\frac{1}{|F|(|F|-1)}\sum_{j\neq c}P(j,c).
\label{eq:globalfps}
\end{equation}
}
The global quantity is straightforward; assigning FPS to a specific family is where the identification problem appears.

\subsection{Why the natural per-family contrast is confounded}
Write the cell mean as
\begin{equation}
P(j,c)=\mu+a_j+b_c+\gamma_{jc},
\label{eq:decomp}
\end{equation}
Here $a_j$ is a judge main effect, $b_c$ a candidate-family main effect, and $\gamma_{jc}$ the judge$\times$candidate interaction. We use the standard double-centered identification constraints $\sum_c b_c=0$, $\sum_c\gamma_{jc}=0$ for every judge $j$, and $\sum_j\gamma_{jc}=0$ for every candidate $c$. Because every judge row has mean 0.5 in the balanced zero-sum design, these constraints set $\mu=0.5$ and absorb the additive judge main effect. Candidate main effects remain and capture average candidate-family strength.

The seemingly natural statistic compares a judge's diagonal cell with the other cells in the same row:
\begin{align}
\mathrm{FPS}^{\text{row}}_f
&=P(f,f)-\frac{1}{3}\sum_{c\neq f}P(f,c)\\
&=\frac{4}{3}(b_f+\gamma_{ff}).
\label{eq:row}
\end{align}
The problem is the $b_f$ term. A strong candidate family can make its same-family judge appear biased even when the interaction is modest, while a weak candidate can conceal a real same-family lift.

To identify the interaction, we keep the candidate family fixed and change the judge:
\begin{align}
\mathrm{FPS}^{\text{col}}_f
&=P(f,f)-\frac{1}{3}\sum_{j\neq f}P(j,f)\\
&=\frac{4}{3}\gamma_{ff}.
\label{eq:col}
\end{align}
This asks the question we actually care about: does candidate family $f$ receive more support from its same-family judge than from judges in the other families?

\begin{figure}[t]
\centering
\includegraphics[width=\columnwidth]{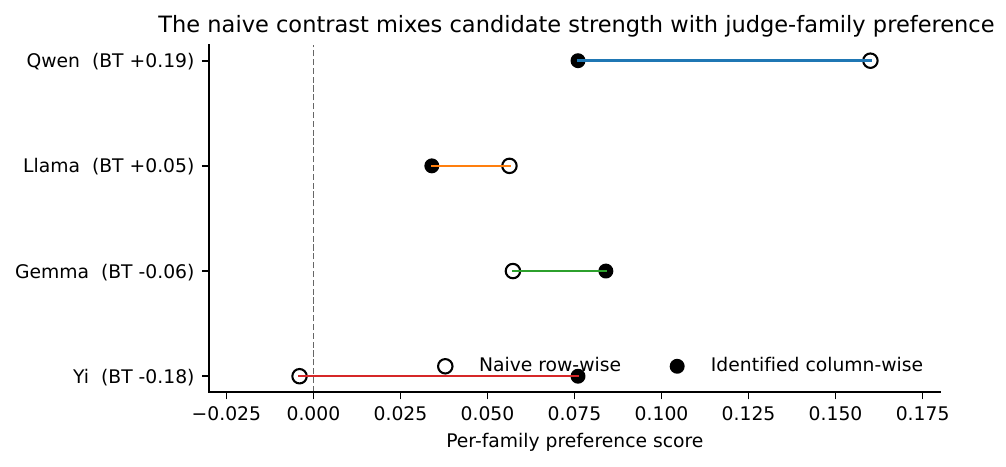}
\caption{\textbf{The same data tell a different story when candidate quality is held fixed.} Open markers show the naive row-wise contrast; filled markers show the identified column-wise contrast. The naive statistic closely tracks Bradley--Terry candidate ability ($r=0.95$). The identified statistic compares judges on the same candidate family and no longer follows ability ($r=-0.24$). Table~\ref{tab:fps} reports inference for the identified estimator.}
\label{fig:estimators}
\end{figure}

Figure~\ref{fig:estimators} shows the practical consequence. Bradley--Terry family abilities are Qwen $+0.190$, Llama $+0.052$, Gemma $-0.060$, and Yi $-0.182$, and the naive row-wise scores track them closely ($r=0.95$): Qwen appears highly biased (0.160) while Yi appears almost unbiased ($-0.004$). After candidate strength is removed, the correlation falls to $r=-0.24$ and all four lifts are positive. Candidate quality adds about $+0.084$ to Qwen's naive score and a roughly $-0.080$ term hides Yi's positive interaction.

\subsection{Three independent checks give the same answer}
Three checks lead to the same correction. Double-centering the $4\times4$ matrix isolates the diagonal interaction terms, and Bradley--Terry residualization gives lifts of 0.025/0.057/0.063/0.057. Prompt-level permutation explains why the naive null is not centered at zero: its per-family means are $+0.023/+0.084/-0.026/-0.080$, matching $\tfrac{4}{3}b_f$ to four decimals. Re-centering by those means recovers the identified estimate and supplies the per-family $p$-values in Table~\ref{tab:fps}, followed by two-sided Benjamini--Hochberg correction.

\section{Results: The Corrected Effect Is Positive in All Four Families}
Global FPS is 0.067, with mean diagonal support of 0.551 versus 0.483 off the diagonal. Every identified family score is also positive and individually significant (Table~\ref{tab:fps}), with a corrected range of 0.034--0.084 that no longer tracks candidate strength. Qwen has the highest BT ability but a middle-sized lift, while Gemma has the largest lift despite being neither the strongest candidate nor a stable judge by position (Section~\ref{sec:position}).

\begin{table}[t]
\centering
\small
\begin{tabular}{lccc}
\toprule
Family & FPS & 95\% CI & $p$ (BH) \\
\midrule
Global & 0.067 & [0.053, 0.084] & 0.0002 \\
Gemma & 0.084 & [0.056, 0.111] & $<0.001$ \\
Yi    & 0.076 & [0.050, 0.103] & $<0.001$ \\
Qwen  & 0.076 & [0.050, 0.102] & $<0.001$ \\
Llama & 0.034 & [0.007, 0.060] & 0.012 \\
\bottomrule
\end{tabular}
\caption{\textbf{Identified Family Preference Score.} Global CI: prompt-cluster bootstrap ($n=2{,}000$); global $p$: prompt-level permutation ($n=5{,}000$). Per-family intervals and two-sided BH-corrected $p$-values use the permutation distribution re-centered on the correct quality-dependent null.}
\label{tab:fps}
\end{table}

\begin{figure}[t]
\centering
\includegraphics[width=\columnwidth]{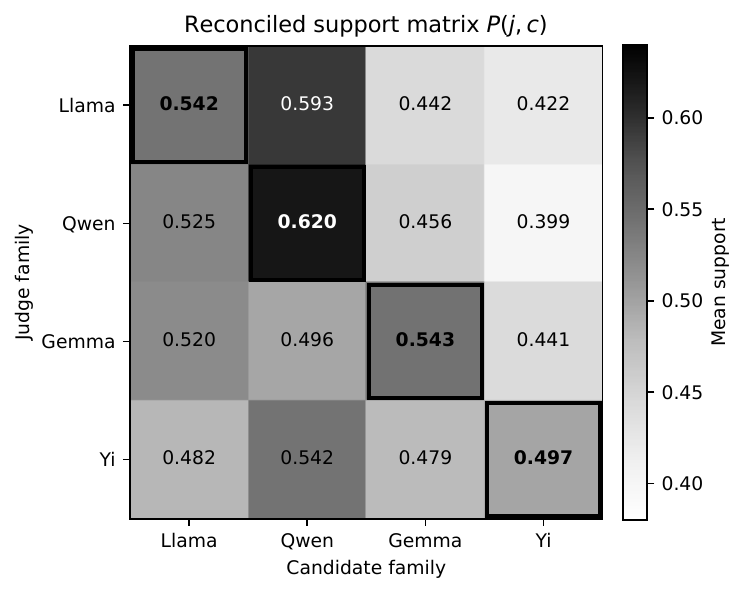}
\caption{\textbf{Reconciled support matrix $P(j,c)$.} Rows are judge families and columns are candidate families. Each row sums to 2.00 by construction. Boxed diagonal cells are the same-family cells. Each diagonal cell is above the mean support given by the other three judges to that same candidate family.}
\label{fig:matrix}
\end{figure}

The full matrix in Figure~\ref{fig:matrix} shows the same pattern directly. Each diagonal cell is higher than the mean support given by the other three judges to that candidate family: Llama 0.542 versus 0.509, Qwen 0.620 versus 0.544, Gemma 0.543 versus 0.459, and Yi 0.497 versus 0.421. Minor differences from printed rounded values arise from rounding before averaging. The corrected pattern is four elevated diagonals, not a single unusually high Qwen cell.

\section{The Effect Survives Measured Quality Controls}
\subsection{Panel-internal controls}
We test whether measured quality and representation features absorb the effect with a sequence of fractional-logit generalized linear models using a Binomial family, logit link, and prompt-clustered sandwich standard errors. The 4,656 reconciled judge--prompt--pair observations are represented as 9,312 candidate-orientation rows. Controls enter the model in blocks (Table~\ref{tab:gee}). A separate quasi-binomial GEE with an exchangeable working correlation is used as a robustness check (Appendix~\ref{app:gee}).

\begin{table}[t]
\centering
\small
\setlength{\tabcolsep}{3.2pt}
\begin{tabular}{clccc}
\toprule
M & Controls added & $\hat\beta$ & OR & $p$ \\
\midrule
1 & same-family only & 0.270 & 1.31 & $4.5\!\times\!10^{-18}$ \\
2 & + BT quality & 0.274 & 1.32 & $7.2\!\times\!10^{-18}$ \\
3 & + style similarity & 0.260 & 1.30 & $1.4\!\times\!10^{-15}$ \\
4 & + length, cand. side & 0.294 & 1.34 & $1.8\!\times\!10^{-16}$ \\
5 & + judge/candidate FE & 0.298 & 1.35 & $4.1\!\times\!10^{-17}$ \\
6 & + opponent, category FE & 0.301 & 1.35 & $1.9\!\times\!10^{-17}$ \\
\bottomrule
\end{tabular}
\caption{\textbf{Nested fractional-logit controls.} The same-family log-odds coefficient remains stable as quality, similarity, length, candidate-side orientation, and family controls are added. Prompt-clustered sandwich SEs are approximately 0.031--0.036.}
\label{tab:gee}
\end{table}

The same-family coefficient does not decline as controls are added, moving from 0.270 in M1 to 0.301 in M6; M5 is 0.298. Because Bradley--Terry quality is panel-derived, we also re-estimate BT after removing all 2,328 same-family pairs; family abilities stay unchanged to four decimals and the M2 coefficient is 0.271. Style similarity from all-mpnet-base-v2 embedding cosine is not predictive ($p=0.91$). This measure captures broad semantic and stylistic similarity, not individual surface features such as markdown structure, list use, or hedging.

\subsection{Independent human quality anchor}
Panel-derived controls can still be circular, so we repeat the analysis using blind human consensus as an external quality signal. Of the 266 consensus items, 248 join to reconciled data from all four primary judges. The resulting matched set contains 992 reconciled judge--pair observations, represented as 1,984 orientation rows across 154 prompt clusters.

On the fixed human-matched subset, the same-family coefficient is 0.370 without the human control (SE 0.059, $p=2.7\times10^{-10}$) and 0.382 with it (SE 0.060, $p=2.0\times10^{-10}$); the human quality coefficient is 0.416 ($p=4.1\times10^{-8}$). The relevant comparison is within this subset, not between 0.370 and the full-sample M1 value of 0.270. An external quality signal therefore does not attenuate the family coefficient.

The human result supports the main finding without proving that all quality confounding is absent. Agreement is noisy ($\kappa=0.482$; 266/400 consensus), so smaller unmeasured differences could remain; the value of this anchor is its independence from the LLM panel.

\section{Likelihood Familiarity as an Explanatory Channel}
\label{sec:familiarity}
One possible explanation for extra same-family support is familiarity. Prior work finds that LLM evaluators can favor text with higher likelihood or lower perplexity \citep{wataoka2024,stureborg2024,oi2024}. Our audit logs judge-side likelihoods, allowing us to test whether the same pattern extends across model families.

\paragraph{Same-family text is more predictable to the judge.}
Same-family responses have a judge-side mean per-token log-probability of $-0.69$, compared with $-1.38$ for cross-family responses. The 0.69 natural-log-unit difference is roughly a twofold ratio in geometric-mean token probability. Family overlap is therefore strongly associated with a directly observed familiarity measure.

\paragraph{Adding likelihood advantage reduces the family coefficient.}
We add z-scored judge-side likelihood advantage, defined as candidate minus opponent, to a fully controlled GEE with BT quality advantage, style advantage, length ratio, and judge/candidate fixed effects. Likelihood coverage is 99.5\% ($n=9{,}264$ orientation rows). Before likelihood is added, the same-family coefficient is 0.299. This is nearly the same as Table~\ref{tab:gee}'s M5 estimate of 0.298.

\begin{table}[t]
\centering
\small
\setlength{\tabcolsep}{3.0pt}
\begin{tabular}{lcccc}
\toprule
Model & $\beta_{same}$ & $p$ & $\beta_{lik}$ & $p$ \\
\midrule
Full controls & 0.299 & $<10^{-17}$ & -- & -- \\
+ likelihood adv. & 0.116 & 0.006 & 0.144 & $7\times10^{-6}$ \\
+ same$\times$lik. & 0.109 & 0.010 & 0.142 & $7\times10^{-6}$ \\
\bottomrule
\end{tabular}
\caption{\textbf{Likelihood-familiarity adjustment.} Adding judge-side likelihood advantage reduces the same-family coefficient by 61\%. The interaction coefficient in the final model is 0.010 ($p=0.001$).}
\label{tab:likelihood}
\end{table}

After likelihood advantage is added, the same-family coefficient drops from 0.299 to 0.116, a 61\% attenuation. Likelihood advantage itself strongly predicts support ($\beta=0.144$, $p=7\times10^{-6}$), while the remaining same-family coefficient is still significant ($p=0.006$). In the interaction model, $\beta_{same}=0.109$ ($p=0.010$), $\beta_{lik}=0.142$ ($p=7\times10^{-6}$), and the same-family$\times$likelihood interaction is small and positive ($\beta=0.010$, $p=0.001$).

\begin{figure}[t]
\centering
\includegraphics[width=\columnwidth]{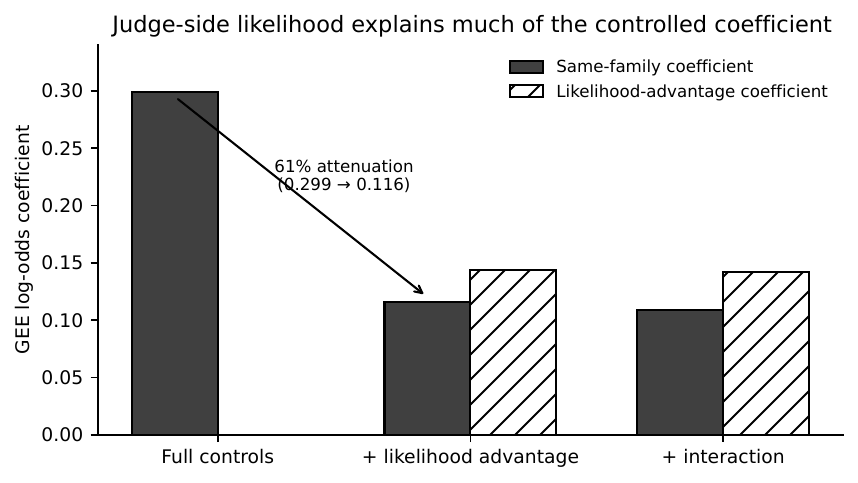}
\caption{\textbf{Coefficient attenuation after adding judge-side likelihood advantage.} The same-family coefficient decreases by 61\%, while likelihood advantage is itself strongly predictive. The 61\% number is descriptive coefficient attenuation across nested nonlinear models, not a causal mediated proportion.}
\label{fig:likelihood}
\end{figure}

\paragraph{Interpretation boundary.}
The 61\% reduction is not a causal share. Non-collapsibility and rescaling can change coefficients across nested nonlinear models, and a common cause could make same-family text both easier to predict and more highly rated. Familiarity may also capture quality that a judge genuinely values, as \citet{chen2025reason} note for self-preference. The supported conclusion is narrower: much of the controlled family coefficient co-varies with a measurable judge--candidate property, and a smaller significant family effect remains after it is included. Judge-side log-probability is therefore a useful audit signal even though its causal role remains unresolved.

\section{Position Bias Is a Separate Failure Mode}
\label{sec:position}
Across all AB/BA presentation pairs, 55.4\% produce different winners. Gemma reverses on 78.8\% of pairs, Yi on 53.4\%, Llama on 47.3\%, and Qwen on 42.1\% ($\chi^2(3)=373.5$, $p<0.001$). Such reversal can be described loosely as judgment noise, but these rates allow a sharper test.

\paragraph{A simple content-noise model has a 50\% ceiling.}
Assume the judge has a stable latent content preference and independently selects that preferred response with probability $p$ in each presentation. Then the AB/BA reversal probability is
\begin{equation}
R_{noise}=2p(1-p)\leq 0.5.
\label{eq:noise}
\end{equation}
The bound also holds for mixtures of different $p$ values across items. Under this simple independent symmetric content-noise model, reversal cannot exceed 50\%.

Pure position-following has a different signature. If a judge chooses the first-shown response with probability $q$, the probability that AB and BA identify different candidate families is
\begin{equation}
R_{slot}=q^2+(1-q)^2,
\label{eq:slot}
\end{equation}
This quantity is above 0.5 whenever $q\neq0.5$ and approaches 1 as first-position following becomes deterministic. Equation~\ref{eq:slot} has symmetric solutions, $q$ and $1-q$; the panel's raw first-position preference (A-rate 0.778) selects the $q>0.5$ branch. Under the pure slot-following model, Gemma's 78.8\% reversal corresponds to $q\approx0.88$, while Yi's 53.4\% corresponds to $q\approx0.63$. Llama and Qwen are below the 50\% bound, so reversal alone cannot tell whether their instability comes from ordinary content noise or milder position bias.

\paragraph{Counterbalancing removes position-following, but attenuates preference.}
Because every pair is judged in both orders, a reversed pair receives $s=0.5$. Pure position-following then contributes no signed preference after reconciliation, which is the intended protection. The same rule also removes any genuine family-preference signal on that reversed pair. High reversal can therefore reduce reconciled FPS even when family-conditioned preference is present.

If we restrict the analysis to position-consistent pairs, global FPS rises to 0.211, roughly three times the headline estimate. We report this as sensitivity evidence only. Consistency is a post-outcome selection that changes both the subset and the judge composition; Qwen contributes many more stable pairs than Gemma. The 0.211 value is therefore neither a corrected estimate nor a formal bound. What is fixed by the design is simpler: reversed pairs contribute exactly zero signed preference to the headline estimator.

This distinction explains the most extreme panel behavior. A Gemma-only judge disagrees with the family-balanced reference on 58.3\% of comparisons, which could look like extreme family bias. Under the position model, however, about 88\% of Gemma's single-order choices follow the first position, leaving little stable content signal in a single-presentation Gemma panel. Position-following and family-conditioned preference are systematic but distinct problems and should be diagnosed separately.

\section{Robustness}
\paragraph{Full-precision judging replication.}
We repeat judging with all four smaller judges loaded at float16 in \texttt{transformers}, removing GGUF quantization from the judging stack. The run uses the original 66-prompt confirmatory allocation and yields 3,168 trials with a 95.5\% parse rate. Candidate responses remain fixed, so the comparison changes only the judging stack. Global FPS is 0.066 (95\% CI [0.045, 0.089]), almost identical to the main estimate of 0.067. All four per-family point estimates remain positive: Gemma 0.115 [0.066, 0.166], Yi 0.083 [0.042, 0.122], Qwen 0.037 [$-0.001$, 0.076], and Llama 0.031 [0.000, 0.062]. Gemma and Yi are individually bounded away from zero at this holdout size; Qwen and Llama replicate in direction but not with individual significance.

\paragraph{Confirmatory holdout and specification multiverse.}
The main-stack confirmatory holdout contains 62 prompts after the response-health filter and gives FPS 0.062 (95\% CI [0.039, 0.086], $p<0.001$). Across 72 Primary-4 specifications varying tie treatment, prompt source, and judge scale, every estimate is positive, ranging from 0.039 to 0.231 with median 0.105; the 0.067 headline lies in the conservative quartile. The multiverse supports the direction of the effect while showing that its magnitude depends on evaluation choices.

\begin{table}[t]
\centering
\small
\setlength{\tabcolsep}{4pt}
\begin{tabular}{lc}
\toprule
Condition & Global FPS \\
\midrule
Headline (rubric, large judges) & 0.067 \\
Full-precision fp16 judges & 0.066 \\
Neutral system prompt & 0.063 \\
MT-Bench + AlpacaEval only & 0.077 \\
WildBench only & 0.039 \\
Small judges (7--9B) & 0.046 \\
Confirmatory holdout & 0.062 \\
Position-consistent pairs only & 0.211 \\
LOO judge (remove Qwen) & 0.037 \\
LOO judge (other removals) & 0.071--0.091 \\
LOO candidate (range) & 0.063--0.102 \\
Multiverse, 72 specs (min/med/max) & 0.039/0.105/0.231 \\
\bottomrule
\end{tabular}
\caption{\textbf{Robustness suite.} All reported global FPS estimates are positive. The range demonstrates directional stability, not a claim that the magnitude is invariant to design choices.}
\label{tab:robust}
\end{table}

No leave-one-out analysis removes the effect, although the families do not contribute equally. Removing Qwen as a judge lowers global FPS from 0.067 to 0.037, the largest single-family change; removing any other judge leaves 0.071--0.091. Removing candidate families gives 0.063--0.102. A coarser family-label placebo shuffle produces a null centered near zero, with the observed FPS above 96.2\% of draws ($p=0.038$). We treat that placebo as supplementary to the prompt-level permutation test ($p=0.0002$), not as a substitute for it.

\section{Consequences for Evaluation Practice}
\paragraph{Panel composition changes decisions.}
Panel composition changes decisions as well as scores. Relative to a family-balanced four-judge reference, a panel where one judge shares a family with one candidate changes 18.5\% of pairwise outcomes (Wilson 95\% CI [16.4, 20.8]). Single-family panels diverge by 22.1\% for Qwen and 58.3\% for Gemma, while removing one family from an otherwise balanced panel changes 2.2--13.4\%. The winner can therefore depend on who judges.

\paragraph{The effect is large relative to common leaderboard gaps.}
The 6.7 pp global lift exceeds 8 of 9 adjacent length-controlled win-rate gaps in the AlpacaEval 2.0 top-10 snapshot retrieved on 2026-05-07, whose median adjacent gap is 1.09 pp \citep{dubois2024}. Because FPS and AlpacaEval win-rate are different quantities, this is only a scale comparison. The observed evaluator-family effect is nevertheless large relative to margins often used to separate neighboring systems.

\paragraph{A minimal audit protocol.}
For a pairwise LLM-as-judge panel, our results motivate five basic checks:
\begin{enumerate}
    \item Use a per-family statistic that identifies the family interaction. In a balanced zero-sum matrix, this can be the column-wise FPS or an equivalent permutation de-centering. The naive row-wise diagonal contrast should not be interpreted as family bias.
    \item Do not let one candidate family also dominate the judge panel.
    \item Judge both response orders and report the AB/BA reversal rate. Under the simple independent content-noise model used here, reversal above 50\% is incompatible with content noise alone and makes single-presentation conclusions difficult to trust.
    \item Record judge-side response likelihoods. A large same-family likelihood gap is a warning that familiarity may be related to verdicts.
    \item Report how outcomes change when panel composition changes, preferably relative to a family-balanced reference.
\end{enumerate}

\section{Conclusion}
Who judges matters, but the effect has to be measured correctly. In our fully crossed open-weight panel, the usual row-wise per-family contrast is strongly confounded with candidate ability and gives a distorted view of family differences. Holding the candidate family fixed produces a simpler result: every tested family receives more support from its same-family judge, with identified lifts of 3.4--8.4 percentage points and global FPS 0.067.

Measured quality differences do not remove the effect: it remains under panel-based controls, an independent human-consensus anchor, and an unquantized float16 judging replication. Judge-side likelihood accounts for much of the statistical association, reducing the controlled coefficient by 61\%, while a smaller significant family effect remains. This is descriptive attenuation, not a causal mechanism estimate; a direct mechanism test would manipulate predictability while holding response quality fixed.

Position creates a second source of evaluator dependence. High AB/BA reversal shows that some judges follow presentation order strongly; counterbalancing removes pure position-following but also suppresses family-preference signal on reversed pairs. Reliable LLM-as-judge evaluation therefore requires provenance-aware measurement alongside explicit judge diagnostics. The prepared reproducibility archive includes the prompts, inference settings, all 9,312 judgments, per-trial judge-side likelihoods, human annotations, and analysis artifacts needed to reproduce the audit.

\section*{Limitations}
The evidence covers four open-weight families at the tested versions, an English open-ended prompt distribution, and one main inference stack. We do not test closed-weight judges or verifiable domains such as code and mathematics, where ground truth can change how apparent self- or family-preference should be interpreted. \citet{chen2025reason} show that some self-preference on verifiable tasks reflects genuinely better outputs, while proof-domain judge audits report different reliability patterns \citep{gonzalez2026}.

The 61\% likelihood result compares coefficients across nested nonlinear models; it is not causal mediation. A stronger mechanism experiment would vary predictability or familiarity while keeping response quality fixed. Our likelihood feature also uses only the first 400 characters of each response. We have not shown that the attenuation percentage is unchanged under other truncation windows, although the prepared likelihood cache permits those checks.

The position-consistent FPS of 0.211 comes from a selected subset with a different mix of trials and judges, so it is sensitivity evidence rather than a corrected estimate or formal bound. Response quality has no perfect gold standard here: standard BT, off-diagonal-only BT, and blind human consensus all leave the effect in place, but unmeasured differences remain possible. The float16 replication also uses the smaller judges on the confirmatory allocation; a full-scale, full-precision replication would give tighter family-level intervals and a stronger quantization test.

\section*{Ethical Considerations}
Human annotation involved three researchers with graduate-level NLP training who were recruited independently of the authors. One annotator was excluded under the pre-specified quality criteria in Appendix~\ref{app:human}; the other two provide the consensus and human-anchor analyses reported in the paper. We collected no demographic data, targeted no vulnerable population, and used no deception.

The audit is dual-use because evaluator provenance could be used to construct a panel that favors one model family. We plan to release the measurements because they also expose that risk and support direct mitigations through balanced panels, order counterbalancing, likelihood logging, and composition-sensitivity reporting. Original annotations, judgments, and analysis artifacts are intended for release under CC BY 4.0, with code under an open-source software license; third-party prompts, model weights, and derived materials remain under their original licenses.

\appendix
\section{Falcon 40B Exclusion and Position Diagnostics}
\label{app:falcon}
Falcon 40B was excluded from the Primary-4 headline under two criteria set before FPS was computed. Its judge tie rate was 88.5\%, versus a 3.4\% mean for retained families, and its mean candidate support was 0.063 versus 0.56--0.65. Candidate support is the mean reconciled support received as a candidate across judges and opponents. Including Falcon would therefore mix a major response-health/quality failure into the preference quantity, so Full-5 results are kept only as sensitivity rows in the specification curve.

AB/BA reversal rates by judge are Gemma 78.8\%, Yi 53.4\%, Llama 47.3\%, and Qwen 42.1\%; the global rate is 55.4\% ($\chi^2(3)=373.49$, $p<0.001$). Under the content-noise model in Section~\ref{sec:position}, reversal is $2p(1-p)\leq0.5$. Under pure first-slot following it is $q^2+(1-q)^2$, which is above 0.5 whenever $q\neq0.5$. Using the observed first-position preference to choose the $q>0.5$ branch gives $q\approx0.88$ for Gemma and $q\approx0.63$ for Yi. Llama and Qwen stay below the 50\% diagnostic bound, so their reversal rates alone cannot separate content noise from mild position bias.

\section{Human Annotation Protocol}
\label{app:human}
Three graduate-level NLP annotators labeled 400 blinded pairwise items as A/B/Tie with 1--5 confidence. One annotator was excluded before any FPS analysis after meeting all three pre-specified criteria: nearly uniform 4--5 confidence, zero notes while each retained annotator left at least 229, and Cohen's $\kappa\approx0.14$--0.18 with both retained annotators, below the 0.40 moderate-agreement threshold in the plan \citep{landis1977}.

The retained annotators have $\kappa=0.482$, 66.5\% exact agreement, and 266 consensus items. On this set, the LLM panel matches 57.1\% of labels versus 42.1\% for the majority-class baseline, a 15.0 pp gain. The fully matched 248-item human-GEE subset has 61.3\% exact panel--human agreement; treating all 134 non-consensus items as ties gives 38.0\% exact match. These labels provide an independent quality control and task-difficulty reference, not a definition or validation of FPS.

\section{Regression Specification and GEE Robustness}
\label{app:gee}
The nested M1--M6 control table uses a fractional-logit generalized linear model with a Binomial family and logit link. Reconciled support in $[0,1]$ is the outcome; 62.3\% of observations are fractional after reconciliation. The model is fit with \texttt{statsmodels.formula.api.glm}, and uncertainty uses sandwich covariance clustered at \texttt{prompt\_id} ($n=194$ clusters). The candidate-side indicator records whether the expanded orientation row corresponds to canonical \texttt{family\_1} or \texttt{family\_2}; it is not the original AB/BA presentation position.

Main nested-model covariates are Bradley--Terry candidate-minus-opponent advantage, judge--candidate style similarity and style advantage, log token-length ratio, the candidate-side indicator, and progressively added judge, candidate, opponent, and category fixed effects. A separate quasi-binomial GEE with a Binomial family, logit link, exchangeable working correlation, and prompt-level sandwich standard errors is used as a robustness check. To avoid the collinearity that arises when Bradley--Terry advantage and candidate/opponent fixed effects are entered together, that GEE uses the quality contrast without the saturated candidate/opponent fixed-effect block. The likelihood models in Section~\ref{sec:familiarity} are also prompt-clustered GEE specifications and add \texttt{logprob\_advz}, the z-scored candidate-minus-opponent difference in judge-side mean per-token log-probability over the first 400 characters. The nested fractional-logit table is the main control ladder; the GEE is a separate robustness analysis.

\section{Full-Precision Replication Detail}
\label{app:fp16}
The full-precision replication loads the four smaller judges (Llama-3.1-8B-Instruct, Qwen2.5-7B-Instruct, Gemma-2-9B-it, and Yi-1.5-9B-Chat) at float16 with \texttt{transformers}, without GGUF or quantization. They re-judge the original 66-prompt confirmatory allocation across all six candidate pairs and both response orders, giving 3,168 trials. This is the pre-filter set; Section~9 uses the 62 prompts remaining after the response-health filter. Candidate responses are unchanged, decoding is greedy, parse rate is 95.5\%, and raw tie rate is 11.5\%. Reconciliation and FPS definitions are unchanged. Global FPS is 0.066 (95\% CI [0.045, 0.089]); family scores are Gemma 0.115 [0.066, 0.166], Yi 0.083 [0.042, 0.122], Qwen 0.037 [$-0.001$, 0.076], and Llama 0.031 [0.000, 0.062].


\begin{thebibliography}{99}

\bibitem[Chen et~al.(2025a)Chen, Wei, Zhu, Feng, and Meng]{chen2025reason}
Wei-Lin Chen, Zhepei Wei, Xinyu Zhu, Shi Feng, and Yu Meng. 2025a.
Do LLM evaluators prefer themselves for a reason?
\textit{arXiv preprint arXiv:2504.03846}.

\bibitem[Chen et~al.(2025b)Chen, Wang, Zhang, Hu, and Lin]{chen2025beyond}
Zhi-Yuan Chen, Hao Wang, Xinyu Zhang, Enrui Hu, and Yankai Lin. 2025b.
Beyond the Surface: Measuring Self-Preference in LLM Judgments.
In \textit{Proceedings of EMNLP 2025}, pages 1653--1672.

\bibitem[Dubois et~al.(2024)Dubois, Galambosi, Liang, and Hashimoto]{dubois2024}
Yann Dubois, Balazs Galambosi, Percy Liang, and Tatsunori B. Hashimoto. 2024.
Length-controlled AlpacaEval: A simple way to debias automatic evaluators.
\textit{arXiv preprint arXiv:2404.04475}.

\bibitem[Goel et~al.(2025)Goel, Struber, Auzina, Chandra, Kumaraguru, Kiela, Prabhu, Bethge, and Geiping]{goel2025}
Shashwat Goel, Joschka Struber, Ilze Amanda Auzina, Karuna K. Chandra, Ponnurangam Kumaraguru, Douwe Kiela, Ameya Prabhu, Matthias Bethge, and Jonas Geiping. 2025.
Great Models Think Alike and this Undermines AI Oversight.
In \textit{Proceedings of the 42nd International Conference on Machine Learning}.

\bibitem[Gonzalez et~al.(2026)]{gonzalez2026}
Santiago Gonzalez et al. 2026.
QEDBENCH: Quantifying the alignment gap in automated evaluation of university-level mathematical proofs.
\textit{arXiv preprint arXiv:2602.20629}.

\bibitem[Kohli(2026)]{kohli2026}
Guneet Kohli. 2026.
Nine judges, two effective votes: Correlated errors undermine LLM evaluation panels.
\textit{arXiv preprint arXiv:2605.29800}.

\bibitem[Landis and Koch(1977)]{landis1977}
J. Richard Landis and Gary G. Koch. 1977.
The measurement of observer agreement for categorical data.
\textit{Biometrics}, 33(1):159--174.

\bibitem[Li et~al.(2026)Li, Sun, Huang, Zhong, Jiang, Han, Zhang, Wang, and Liu]{li2026}
Dawei Li, Renliang Sun, Yue Huang, Ming Zhong, Bohan Jiang, Jiawei Han, Xiangliang Zhang, Wei Wang, and Huan Liu. 2026.
Preference Leakage: A Contamination Problem in LLM-as-a-Judge.
In \textit{International Conference on Learning Representations}.

\bibitem[Lin et~al.(2025)Lin, Deng, Chandu, Brahman, Ravichander, Pyatkin, Dziri, Le Bras, and Choi]{lin2025}
Bill Yuchen Lin, Yuntian Deng, Khyathi Chandu, Faeze Brahman, Abhilasha Ravichander, Valentina Pyatkin, Nouha Dziri, Ronan Le Bras, and Yejin Choi. 2025.
WildBench: Benchmarking LLMs with challenging tasks from real users in the wild.
In \textit{International Conference on Learning Representations}.

\bibitem[Liu et~al.(2024)Liu, Moosavi, and Lin]{liu2024}
Yiqi Liu, Nafise Moosavi, and Chenghua Lin. 2024.
LLMs as narcissistic evaluators: When ego inflates evaluation scores.
In \textit{Findings of ACL 2024}, pages 12688--12701.

\bibitem[Lushtaku et~al.(2026)Lushtaku, Kargi, Elganzory, Ferreira, Salamanca, Kreutzer, and Salinas]{lushtaku2026}
Erlis Lushtaku, Bora Kargi, Ali Elganzory, Fabio Ferreira, Alejandro R. Salamanca, Julia Kreutzer, and David Salinas. 2026.
JudgeArena: A Unified Framework for Reproducible LLM-Judge Evaluation.
\textit{arXiv preprint arXiv:2608.02620}.

\bibitem[Oi et~al.(2024)Oi, Kaneko, Koike, Loem, and Okazaki]{oi2024}
Masanari Oi, Masahiro Kaneko, Ryuto Koike, Mengsay Loem, and Naoaki Okazaki. 2024.
Likelihood-based Mitigation of Evaluation Bias in Large Language Models.
In \textit{Findings of ACL 2024}, pages 3237--3245.

\bibitem[Panickssery et~al.(2024)Panickssery, Bowman, and Feng]{panickssery2024}
Arjun Panickssery, Samuel R. Bowman, and Shi Feng. 2024.
LLM evaluators recognize and favor their own generations.
In \textit{Advances in Neural Information Processing Systems}, volume 37.

\bibitem[Pombal et~al.(2026)Pombal, Rei, and Martins]{pombal2026}
Jose Pombal, Ricardo Rei, and Andre F. T. Martins. 2026.
Self-preference bias in rubric-based evaluation of large language models.
\textit{arXiv preprint arXiv:2604.06996}.

\bibitem[Shi et~al.(2025)Shi, Ma, Liang, Diao, Ma, and Vosoughi]{shi2025}
Lin Shi, Chiyu Ma, Wenhua Liang, Xingjian Diao, Weicheng Ma, and Soroush Vosoughi. 2025.
Judging the Judges: A Systematic Study of Position Bias in LLM-as-a-Judge.
In \textit{Proceedings of IJCNLP-AACL}, pages 292--314.

\bibitem[Soumik(2026)]{soumik2026}
Sadman Kabir Soumik. 2026.
Judging the Judges: A Systematic Evaluation of Bias Mitigation Strategies in LLM-as-a-Judge Pipelines.
\textit{Transactions on Machine Learning Research}.

\bibitem[Spiliopoulou et~al.(2025)Spiliopoulou, Fogliato, Burnsky, Soliman, Ma, Horwood, and Ballesteros]{spiliopoulou2025}
Evangelia Spiliopoulou, Riccardo Fogliato, Hanna Burnsky, Tamer Soliman, Jie Ma, Graham Horwood, and Miguel Ballesteros. 2025.
Play Favorites: A Statistical Method to Measure Self-Bias in LLM-as-a-Judge.
\textit{arXiv preprint arXiv:2508.06709}.

\bibitem[Stureborg et~al.(2024)Stureborg, Alikaniotis, and Suhara]{stureborg2024}
Rickard Stureborg, Dimitris Alikaniotis, and Yoshi Suhara. 2024.
Large language models are inconsistent and biased evaluators.
\textit{arXiv preprint arXiv:2405.01724}.

\bibitem[Usami et~al.(2026)Usami, Hara, Tsuboi, and Matsuda]{usami2026}
Hiroyasu Usami, Keisuke Hara, Ayato Tsuboi, and Naohiko Matsuda. 2026.
LLM Judges Have Dark Current: A Psychometric Datasheet for LLM-as-a-Judge Evaluation.
\textit{arXiv preprint arXiv:2606.15610}.

\bibitem[Verga et~al.(2024)Verga, Hofstatter, Althammer, Su, Piktus, Arkhangorodsky, Xu, White, and Lewis]{verga2024}
Pat Verga, Sebastian Hofstatter, Sophia Althammer, Yixuan Su, Aleksandra Piktus, Arkady Arkhangorodsky, Minjie Xu, Naomi White, and Patrick Lewis. 2024.
Replacing Judges with Juries: Evaluating LLM Generations with a Panel of Diverse Models.
\textit{arXiv preprint arXiv:2404.18796}.

\bibitem[Wang et~al.(2024)Wang, Li, Chen, Cai, Zhu, Lin, Cao, Kong, Liu, Liu, and Sui]{wang2024}
Peiyi Wang, Lei Li, Liang Chen, Zefan Cai, Dawei Zhu, Binghuai Lin, Yunbo Cao, Lingpeng Kong, Qi Liu, Tianyu Liu, and Zhifang Sui. 2024.
Large language models are not fair evaluators.
In \textit{Proceedings of ACL 2024}, pages 9440--9450.

\bibitem[Wataoka et~al.(2024)Wataoka, Takahashi, and Ri]{wataoka2024}
Koki Wataoka, Tsubasa Takahashi, and Ryokan Ri. 2024.
Self-preference bias in LLM-as-a-Judge.
\textit{arXiv preprint arXiv:2410.21819}.

\bibitem[Yang et~al.(2026a)Yang, Hu, Qiu, Deng, Jiao, and Zhou]{yang2026a}
Jinming Yang, Zheng Hu, Chuxian Qiu, Zhenyu Deng, Xinshan Jiao, and Tao Zhou. 2026a.
Quantifying and mitigating self-preference bias of LLM judges.
\textit{arXiv preprint arXiv:2604.22891}.

\bibitem[Yang et~al.(2026b)Yang, Hou, and Yang]{yang2026b}
Zongyou Yang, Yinghan Hou, and Xiaokun Yang. 2026b.
When the Judge Changes, So Does the Measurement: Auditing LLM-as-Judge Reliability.
\textit{arXiv preprint arXiv:2607.08535}.

\bibitem[Zheng et~al.(2023)Zheng, Chiang, Sheng, Zhuang, Wu, Zhuang, Lin, Li, Li, Xing, Zhang, Gonzalez, and Stoica]{zheng2023}
Lianmin Zheng, Wei-Lin Chiang, Ying Sheng, Siyuan Zhuang, Zhanghao Wu, Yonghao Zhuang, Zi Lin, Zhuohan Li, Dacheng Li, Eric P. Xing, Hao Zhang, Joseph E. Gonzalez, and Ion Stoica. 2023.
Judging LLM-as-a-Judge with MT-Bench and Chatbot Arena.
In \textit{Advances in Neural Information Processing Systems}, volume 36.

\end{thebibliography}
\end{document}